\documentclass[runningheads]{llncs}

\usepackage{eccv}

\usepackage{eccvabbrv}

\usepackage{graphicx}
\usepackage{booktabs}
\usepackage{multirow}
\usepackage{colortbl}
\usepackage{tabularx}
\usepackage[table]{xcolor}
\usepackage{makecell}
\usepackage{amssymb}  
\usepackage{pifont}   
\usepackage{bm}
\usepackage{wrapfig}
\usepackage{array}
\usepackage[accsupp]{axessibility}  

\usepackage{hyperref}

\usepackage{orcidlink}

\begin{document}

\title{EchoVLA: Robotic Vision-Language-Action Model with Synergistic Declarative Memory for Mobile Manipulation}

\titlerunning{EchoVLA}

\author{Min Lin\inst{1} \and
Xiwen Liang\inst{1} \and
Bingqian Lin\inst{2} \and
Jingzhi Liu\inst{1} \and 
Zijian Jiao\inst{1} \and
Kehan Li\inst{1} \and
Ziang Yan\inst{1} \and 
Yu Sun\inst{1} \and
Weijia Liufu\inst{1} \and
Yuhan Ma\inst{1} \and
Jiarui Hu\inst{1} \and 
Yuecheng Liu\inst{3} \and
Shen Zhao\inst{1} \and
Yuzheng Zhuang\inst{3} \and
Xiaodan Liang\inst{1,4}\thanks{Corresponding author.}}

\authorrunning{M.~Lin et al.}

\institute{Shenzhen Campus of Sun Yat-sen University, Shenzhen, China \and
Shanghai Jiao Tong University, Shanghai, China \and
Huawei Noah's Ark Lab, China \and
Yinwang Intelligent Technology Co., Ltd., China\\ 
\email{linm57@mail2.sysu.edu.cn, xdliang32@gmail.com}}
\maketitle


\begin{abstract}
Recent progress in Vision–Language–Action (VLA) models has enabled embodied agents to interpret multimodal instructions and perform complex tasks.
However, existing VLAs are mostly confined to short-horizon, table-top manipulation, lacking the memory and reasoning capability required for mobile manipulation, where agents must coordinate navigation and manipulation under changing spatial contexts.
In this work, we present EchoVLA, a memory-aware VLA model for mobile manipulation.
EchoVLA incorporates a synergistic declarative memory inspired by the human brain, consisting of a scene memory that maintains a collection of spatial–semantic maps and an episodic memory that stores task-level experiences with multimodal contextual features.
The two memories are individually stored, updated, and retrieved based on current observations, task history, and instructions, and their retrieved representations are fused via coarse- and fine-grained attention to guide base–arm diffusion policies.
To support large-scale training, we further introduce MoMani, an automated benchmark that generates expert-level trajectories through multimodal large language model (MLLM)–guided planning and feedback-driven refinement, supplemented with real-robot demonstrations.
Comprehensive simulated and real-world results demonstrate that EchoVLA substantially improves overall performance, e.g., it achieves the highest success rates of 0.52 on manipulation/navigation tasks and 0.31 on mobile manipulation tasks in simulation, exceeding the strong baseline $\pi_{0.5}$ by +0.20 and +0.11, respectively.
\end{abstract}

\section{Introduction}
\label{sec:intro}

\begin{figure}[htbp]
    \centering
    \includegraphics[width=0.9\textwidth]{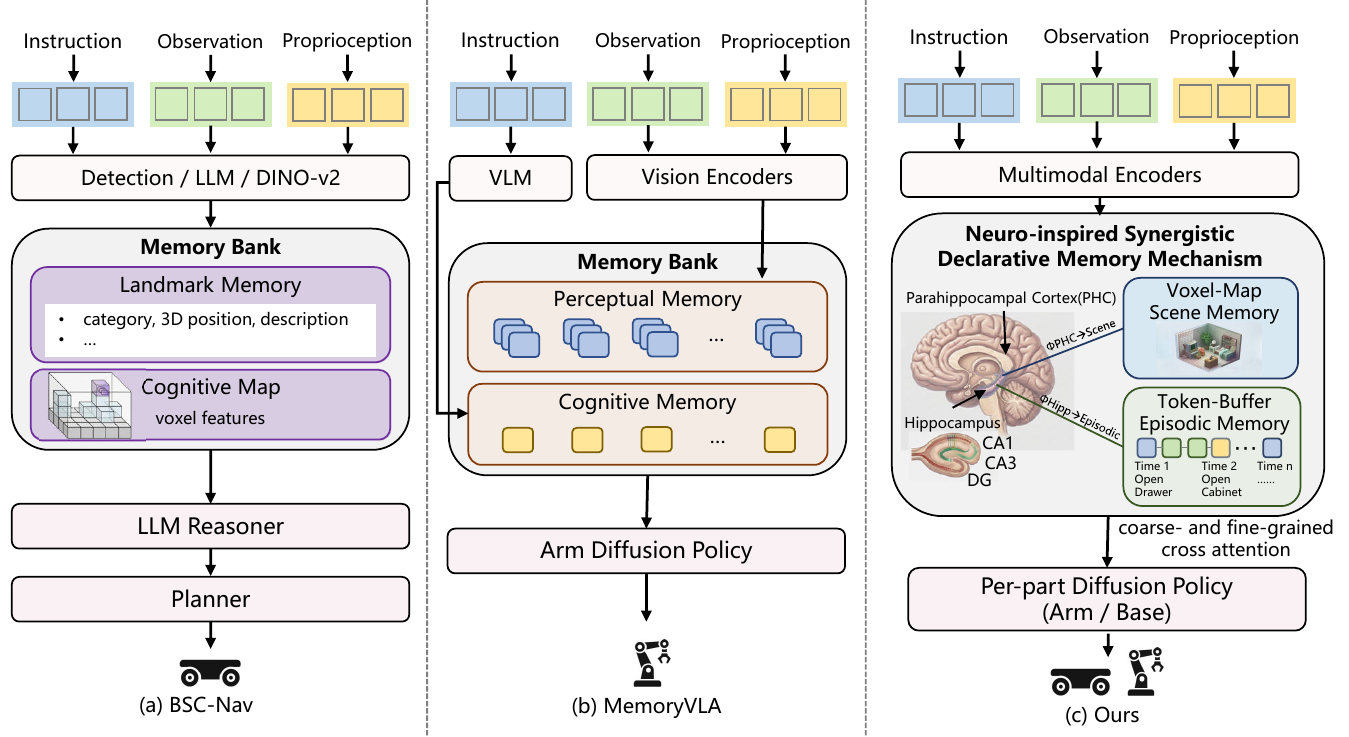}
    \caption{Comparison of memory designs for mobile manipulation control. (a) BSC-Nav \cite{ruan2025reactive} uses a memory bank with landmark memory and a cognitive map to support LLM-based reasoning and planning. (b) MemoryVLA \cite{shi2025memoryvla} builds a memory bank with perceptual memory and cognitive memory to condition a diffusion policy. (c) EchoVLA (ours) introduces dual memories: a persistent scene memory (voxel map) and a time-indexed episodic memory (token buffer), fused via coarse- and fine-grained cross attention for context-aware per-part diffusion control (arm/base).}
    \label{fig:compare_framework}
\end{figure}

Recent advances in Vision–Language–Action (VLA) models \cite{shridhar2022cliport,reed2022generalist,brohan2022rt,jiang2022vima,zhao2023learning,huang2023voxposer,team2024octo,bjorck2025gr00t,ding2025humanoid} have shown strong potential for general-purpose robot learning, enabling embodied agents to interpret multimodal inputs and perform diverse manipulation tasks.
Models such as RT-2 \cite{zitkovich2023rt}, OpenVLA \cite{kim2024openvla}, and ManipLLM \cite{li2024manipllm} demonstrate that large-scale vision–language pretraining can be extended to robot control, achieving impressive zero-shot generalization across unseen objects and scenes.
However, most existing VLAs are restricted to short-horizon, table-top manipulation, and rely on Markovian control, where each decision depends solely on the current observation.
This limitation prevents consistent reasoning over extended task sequences and hinders long-term spatial understanding.

To overcome this limitation, we draw inspiration from the  declarative memory system in the human brain, 
which comprises complementary subsystems for spatial and experiential memory encoding, respectively.
The  parahippocampal cortex (PHC) and related retrosplenial regions represent the spatial and semantic structure of scenes, providing contextual information about environmental layouts and object relations \cite{epstein1998cortical,epstein2008parahippocampal,aminoff2013role}.
These regions project to the hippocampus, which integrates contextual inputs into temporally organized episodic traces that capture individual experiences and task outcomes \cite{ranganath2012two,eichenbaum2017integration}.
Such a dual-system interaction allows humans to remember where things are and how tasks were accomplished—offering a useful analogy for memory-guided embodied reasoning.

Building on this insight, we propose \textbf{EchoVLA}, a memory-aware Vision--Language--Action model for mobile manipulation.
EchoVLA introduces two complementary memories: a \textbf{scene memory} parallels the PHC by  maintaining a persistent spatial representation as a voxel map, and an \textbf{episodic memory} mirrors hippocampal processing by storing time-indexed multimodal tokens to track task progress.
As shown in Figure~\ref{fig:compare_framework}, 
prior work such as BSC-Nav~\cite{ruan2025reactive} organizes memory as landmark descriptors and a cognitive map for LLM-based reasoning and planning, while MemoryVLA~\cite{shi2025memoryvla} builds perceptual and cognitive memories to condition diffusion control.
In contrast, EchoVLA retrieves from scene memory and episodic memory with separate coarse- and fine-grained cross attention, and fuses them to guide per-part diffusion policies (arm/base). 
This neuro-inspired dual-memory with attention-guided formulation helps maintain coherent spatial context and consistent task execution over long horizons.

We further introduce MoMani, an automated benchmark for mobile manipulation that generates expert-level trajectories via MLLM-guided planning and feedback-based refinement. Compared with existing mobile manipulation benchmarks—which remain limited in scale and task diversity—as well as datasets like RoboTwin 2.0 \cite{chen2025robotwin} and RoboCasa \cite{nasiriany2024robocasa}, MoMani offers richer task coverage and collects real-robot demonstrations on a holonomic mobile platform \cite{wu2024tidybot++} to support large scale training. 

Both simulation and real-world validation reveal that our EchoVLA maintains strong robustness in long-horizon mobile manipulation tasks by leveraging its synergistic declarative memory. On the RoboCasa~\cite{nasiriany2024robocasa} simulator, EchoVLA brings significant performance gain in SR  over strong baseline $\pi_{0.5}$~\cite{pi052025} on manipulation/navigation tasks and mobile manipulation tasks by 0.20 and 0.11, respectively. Real-world experiments on the TidyBot++ platform~\cite{wu2024tidybot++} across a $7\text{m} \times 7\text{m}$ arena shows that 
EchoVLA achieves the highest SR of 0.44, 
outperforming $\pi_{0.5}$~\cite{pi052025} (0.33) and Diffusion Policy~\cite{chi2023diffusion} (0.32).


Our contributions are summarized as follows:
\begin{itemize}
    \item We propose EchoVLA, a neuro-inspired memory-aware VLA model equipped with synergistic scene and episodic memory for mobile manipulation.
    \item We introduce MoMani, an automated benchmark that provides expert-level multimodal trajectories with real-robot demonstrations for scalable embodied data generation. 
    \item  Extensive simulated and real-world results  demonstrate that EchoVLA consistently outperforms strong baselines on various mobile manipulation tasks.
\end{itemize}

\section{Related Work}
\label{sec:related_work}

\subsection{Vision-Language-Action Model}
Driven by advances in visual–language foundation models \cite{radford2021learning,caron2021emerging,liu2024grounding} and large-scale robot datasets \cite{o2024open,bu2025agibot}, Vision–Language–Action (VLA) models unify perception, reasoning, and control within a single multimodal policy.
Representative works such as RT-2 \cite{zitkovich2023rt}, OpenVLA \cite{kim2024openvla}, and ManipLLM \cite{li2024manipllm} discretize continuous actions into token sequences and perform autoregressive prediction, enabling large-scale pretraining but limiting motion continuity.
Recent approaches (e.g., CogACT \cite{li2024cogact}, DexVLA \cite{wen2025dexvla}, HybridVLA \cite{liu2025hybridvla}) introduce diffusion- or regression-based action heads to capture the multimodal distribution of robot trajectories, while AC-DiT \cite{chen2025ac} explicitly models coordination between the mobile base and the manipulator through adaptive diffusion.
Notable works $\boldsymbol{\pi}_{0}$ and $\boldsymbol{\pi}_{0.5}$ \cite{pi02024,pi052025} address hierarchical trajectory diffusion for mobile manipulation. $\boldsymbol{\pi}_{0}$ models the full robot action space, while $\boldsymbol{\pi}_{0.5}$ introduces per-part decomposition for mobile base and manipulator, adding partial episodic memory. Despite improved trajectory fidelity, both lack explicit scene-level memory for planning.
Meanwhile, AnywhereVLA \cite{anywherevla2025} adopts a modular design that combines classical SLAM with a fine-tuned VLA head for robust real-world pick-and-place, highlighting the practicality of hybrid systems.
However, most existing VLAs remain Markovian, relying only on current observations without structured memory for reasoning.


In this work, we take inspiration from the memory system of human brain and propose EchoVLA, which constructs a synergistic declarative memory composed of scene and episodic memories. The scene memory explicitly builds a spatial–semantic map of object layouts and environmental structure, while the episodic memory records task-specific experience traces for contextual retrieval. This dual-memory mechanism enables reasoning in mobile-manipulation tasks beyond the capability of prior single-memory VLAs. A concurrent work MemoryVLA \cite{shi2025memoryvla} augments VLAs with a perceptual–cognitive memory that caches visual features to enhance manipulation stability. However, its memory remains an implicit perceptual cache without an explicit spatial representation or task-level experience storage, limiting its ability for long-horizon task reasoning. 

\subsection{Mobile Manipulation Benchmark}
Early embodied benchmarks such as RoboTHOR \cite{deitke2020robothor} for sim-to-real indoor navigation and iGibson 2.0 \cite{li2021igibson} for object-centric household tasks mainly offer mobile perception and basic interaction. Habitat 2.0 \cite{szot2021habitat} further supports interactive rearrangement with articulated objects, but the mobile–manipulation coupling is still limited to predefined skills. Large-scale suites like BEHAVIOR-100 / 1K \cite{srivastava2022behavior,li2024behavior} and ManiSkill2 \cite{gu2023maniskill2} broaden the task families to 100–1000 household activities or 20+ manipulation families, yet most tasks assume a fixed or simplified manipulation setting rather than full mobile manipulation.
More recent, robotics-oriented simulators move closer to real mobile manipulation: RoboCasa \cite{nasiriany2024robocasa} targets realistic kitchen scenes with 120+ layouts and 2,500+ assets but the task spectrum is still mostly kitchen rearrangement; MoMa-Kitchen \cite{zhang2025moma} provides 100K+ affordance-grounded samples for “last-mile” base positioning but is domain-specific to kitchens; RoboTwin 2.0 \cite{chen2025robotwin} automates data generation for 50 dual-arm tasks in 731-object settings but does not involve base navigation; TidyBot \cite{wu2023tidybot} shows LLM-guided, user-conditioned tidy-up on a mobile manipulator, while TidyBot++ \cite{wu2024tidybot++} mainly contributes an open-source holonomic mobile-manipulation platform for data collection rather than a broad, automatic task generator.
Different from existing benchmarks, our constructed MoMani provides a unified, automated pipeline that produces expert-level trajectories for diverse mobile-manipulation tasks and augments them with real-robot demonstrations.

\begin{figure*}
    \centering
    \includegraphics[width=0.95\textwidth]{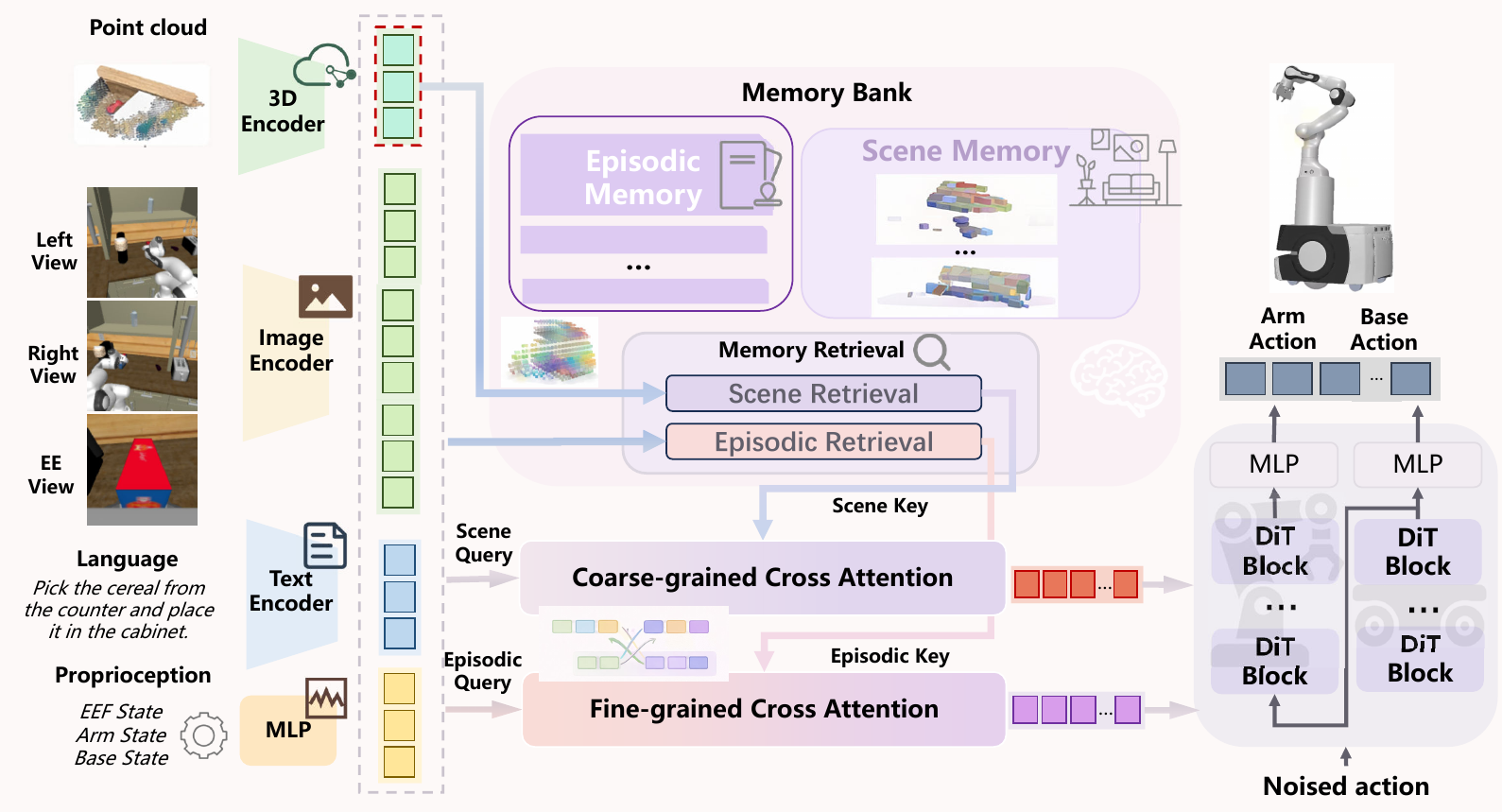}
    \caption{Overview of EchoVLA. Multi-modal observations (RGB views, point clouds, language, and proprioception) are encoded into a unified token sequence. The model retrieves relevant information from both episodic and scene memory through coarse- and fine-grained cross-attention. The retrieved memory augments the diffusion policy, which generates base and arm actions through a per-part denoising process.}
    \label{fig:framework}
\end{figure*}

\newcommand{\zijian}[1]{(\textit{\textcolor{red}{zijian : #1}})}  
\section{EchoVLA}
\label{sec:method}

EchoVLA is a memory-augmented vision–language–action framework designed for mobile manipulation. Its key idea is to maintain two complementary memory systems—scene memory and episodic memory—and integrate them into a coarse-to-fine retrieval hierarchy that improves spatial consistency, temporal reasoning, and manipulation precision.
The overall architecture as illustrated in Figure \ref{fig:framework} consists of:
(1) multimodal state representation (Sec.~\ref{Multimodal State Representation}),
(2) memory retrieval and interaction (Sec.~\ref{Memory Retrieval and Interaction}), and
(3) diffusion-based action generation (Sec.~\ref{Diffusion-based Action Generation}).
We first present the problem formulation of the mobile manipulation task in Sec.~\ref{Problem Formulation}, then we describe each component of EchoVLA in detail.

\subsection{Problem Formulation}
\label{Problem Formulation}
At timestep $t$, the agent observes multi-view RGB-D images $\mathcal{O}_t$, 
proprioceptive states $\mathbf{s}_t$, and a natural-language instruction $\mathcal{I}$.
The goal is to output continuous arm and base actions:
\begin{equation}
(a_t^{\text{arm}},\, a_t^{\text{base}}) = \pi_\theta(\mathcal{I},\, \mathcal{O}_{1:t},\, \mathbf{s}_{1:t}).
\end{equation}

We focus on non-Markovian tasks---
two visually similar frames may correspond to 
completely different progress states (e.g., ``cabinet opened'' vs.\ ``about to open'').
Thus, a memory mechanism is needed.

\subsection{Multimodal State Representation}
\label{sec:multimodal_state_representation}
EchoVLA encodes language, appearance, 3D structure, and robot configuration into a unified token sequence to drive memory retrieval and action generation. 
Natural language instructions are encoded via the text tower of a frozen SigLIP model~\cite{zhai2023sigmoidlosslanguageimage} to produce language tokens $\mathbf{L}$. 
Concurrently, multi-view RGB images from three fixed cameras are processed independently by the frozen SigLIP vision tower to yield per-view features $\mathbf{V}^{(i)}_{t}$, which are concatenated and projected into a shared embedding space as $\mathbf{V}_{t}$ to maintain cross-modal alignment.

Depth observations are fused into a point cloud and encoded by a trainable PointAttn backbone~\cite{wang2022pointattn}. 
The resulting tokens $\mathbf{P}_{t}$ provide fine-grained geometric cues regarding free space and object boundaries while adapting to the specific robot embodiment. 
Lastly, the proprioceptive state $\mathbf{s}_{t}$ is transformed into configuration tokens $\mathbf{R}_{t}$ using a small MLP. 
We then form the unified token sequence:
\begin{equation}
    \mathbf{S}_{t} = [\mathbf{L},\ \mathbf{V}_{t},\ \mathbf{P}_{t},\ \mathbf{R}_{t}],
\end{equation}
which serves as the query for hierarchical memory retrieval and conditions the diffusion policy.

\subsection{Memory Retrieval and Interaction}
\label{Memory Retrieval and Interaction}
EchoVLA maintains two complementary memory banks—scene memory and episodic memory—and retrieves information from them using a two-level hierarchical (coarse-to-fine) attention mechanism. The hierarchy comes from the semantic granularity of the two memories: scene memory provides slowly varying spatial structure, while episodic memory captures fine-grained, time-indexed task progress. The retrieved memory features are later fused with the current tokens and used to condition the diffusion policy.

\begin{figure}[h] 
    \centering
    \includegraphics[width=0.75\textwidth]{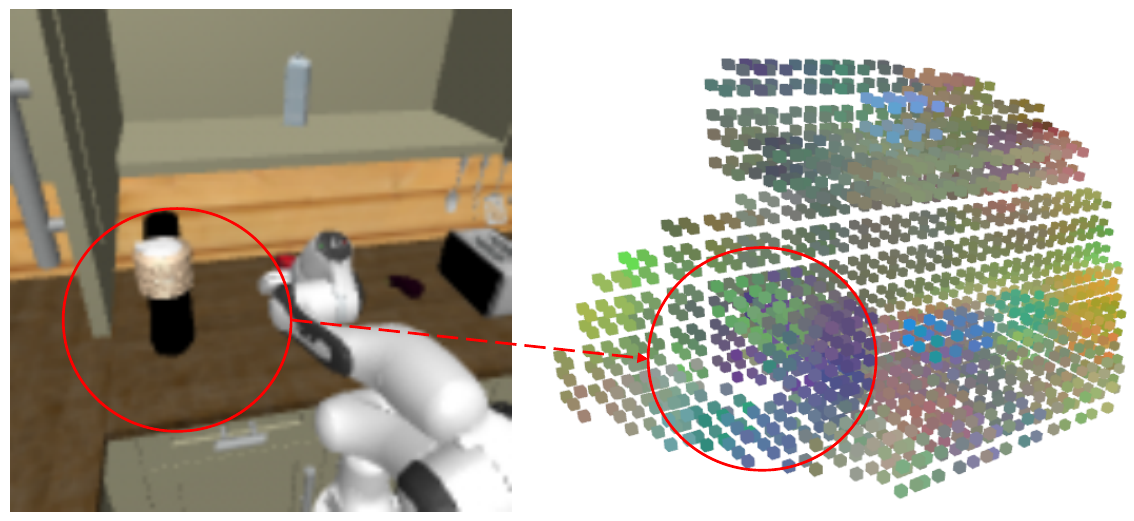} 
    \caption{Visualization of the voxelized 3D feature map, which encodes local geometric structure from depth observations.}
    \label{fig:visualize_map}
\end{figure}
\subsubsection{Scene Memory}

Scene memory $\mathcal{M}^{\mathrm{scene}}$ is maintained as a Key-Value bank, where Keys are 3D spatial indices $(x, y, z)$ and Values are aggregated PointAttn features. It represents a persistent, environment-specific 3D structure refined as new episodes unfold. 
At the beginning of training in a new environment, $\mathcal{M}^{\mathrm{scene}}$ is initialized as an empty voxel grid. As the agent interacts with the environment, depth observations are encoded by the PointAttn network, producing a local 3D feature volume $\mathbf{V}^{3D}_{t} \in \mathbb{R}^{X \times Y \times Z \times C}$ (see Figure~\ref{fig:visualize_map}). While $\mathcal{M}^{\mathrm{scene}}$ stores a global environment-specific representation, feature retrieval for the downstream policy is executed via a spatial-query process, which dynamically selects the top-$k$ relevant entries within the current local camera frustum to maintain efficiency.

To incorporate new observations into the global structure while avoiding redundant updates, a discrepancy-driven rule is formally defined by a reconstruction error $\delta$:
\begin{equation}
    \delta = \|\mathbf{V}_t^{3D} - \mathcal{D}(\mathcal{M}^{\mathrm{scene}}[x,y,z])\|^2,
\end{equation}
where $\mathcal{D}$ is an MLP decoder. Regions whose reconstruction error exceeds a threshold ($\delta > \tau$, where $\tau=0.5$) are updated via exponential moving average (EMA):
\begin{equation}
    \mathcal{M}^{\mathrm{scene}}_{\mathrm{new}} = (1-\alpha)\mathcal{M}^{\mathrm{scene}}_{\mathrm{old}} + \alpha \mathbf{V}_t^{3D},
\end{equation}
with $\alpha=0.2$, while unchanged areas retain their previous values. This enables $\mathcal{M}^{\mathrm{scene}}$ to gradually converge to a stable representation of the environment's geometry.

\subsubsection{Episodic Memory}
Episodic memory stores a short horizon of recent token sequences, each associated with its
timestep. Formally, we maintain a time-indexed buffer:
\begin{equation}
    \mathcal{M}^{\mathrm{epi}}
    = \{(\mathbf{S}_{t-k},\, t-k),\, \ldots,\, (\mathbf{S}_{t-1},\, t-1)\},
    \label{eq:episodic_memory}
\end{equation}
where the window size $k$ is determined by the memory capacity.

Unlike scene memory, which evolves slowly and captures stable 3D structure, episodic memory
preserves fine-grained temporal information tied to recent interactions—such as whether a drawer
has been opened, whether an object has already been grasped, or the precise end-effector
configuration in the immediate past. The memory is maintained as a fixed-size FIFO buffer updated
at each timestep, ensuring that temporally coherent information is retained while avoiding
unbounded growth. By storing the original encoded tokens instead of compressing them into
abstract summaries, the model preserves detailed temporal cues that are essential for resolving
non-Markov ambiguities and maintaining consistent task execution across visually similar states.

\subsubsection{Memory Matching and Attention}
EchoVLA accesses its two memory banks in two steps: it first matches the current representations to
memory entries using cosine similarity, and then interacts with the selected entries via
cross-attention.

For scene memory, the query is the current voxelized 3D feature map
$\mathbf{V}^{3D}_{t}$. We compute cosine similarity between $\mathbf{V}^{3D}_{t}$ and stored scene
maps in $\mathcal{M}^{\mathrm{scene}}$, select the top-$k$ matches
$\mathcal{M}^{\mathrm{scene}}_{\mathrm{sel}}$, and apply coarse cross-attention:
\begin{equation}
    \mathbf{Z}^{\mathrm{scene}}_{t}
    = \mathrm{CrossAttn}
    \bigl(
        \text{q} = \mathbf{V}^{3D}_{t},\;
        \text{k / v}   = \mathcal{M}^{\mathrm{scene}}_{\mathrm{sel}}
    \bigr).
\end{equation}

For episodic memory, the query is the current multi-modal state tokens $\mathbf{S}_{t}$. We
match $\mathbf{S}_{t}$ to historical states in $\mathcal{M}^{\mathrm{epi}}$, select the most
relevant subset $\mathcal{M}^{\mathrm{epi}}_{\mathrm{sel}}$, and apply fine cross-attention:
\begin{equation}
    \mathbf{Z}^{\mathrm{epi}}_{t}
    = \mathrm{CrossAttn}
    \bigl(
        \text{q} = \mathbf{S}_{t},\;
        \text{k / v}   = \mathcal{M}^{\mathrm{epi}}_{\mathrm{sel}}
    \bigr).
\end{equation}

The outputs of the two interactions are combined into a memory-augmented representation:
\begin{equation}
    \mathbf{H}_{t}
    = 
    \bigl[
        \mathbf{Z}^{\mathrm{scene}}_{t},\;
        \mathbf{Z}^{\mathrm{epi}}_{t}
    \bigr],
\end{equation}
which serves as the conditioning input for the diffusion-based action policy.

\subsection{Diffusion-based Action Generation}
\label{Diffusion-based Action Generation}
To model the heterogeneous dynamics of mobile-base motion and arm manipulation, we propose a per-part diffusion policy conditioned on the memory-augmented representation $\mathbf{H}_{t}$.

Each action subspace $p \in \{\mathrm{base},\, \mathrm{arm}\}$ is generated by an independent
denoising diffusion process.
For part $p$, the denoiser predicts the noise of a perturbed action sample:
\begin{equation}
\epsilon_\theta^{(p)} = \text{Denoiser}_p(\mathbf{z}_t, \mathbf{H}_t, t), \quad p \in \{\text{base}, \text{arm}\}
\end{equation}
where $\mathbf{z}_t$ is the noisy action at diffusion step $t$, $\mathbf{H}_t$ captures the fused current observation with retrieved scene-level and episodic context.

During training, the objective minimizes the denoising loss for each part:
\begin{equation}
\mathcal{L} = \sum_{p \in \{\text{base}, \text{arm}\}} \mathbb{E}_{t,\mathbf{z}_t} \Big[\|\epsilon - \epsilon_\theta^{(p)}(\mathbf{z}_t, \mathbf{H}_t, t)\|^2\Big].
\end{equation}

This per-part design enables structured action generation, allowing coordinated yet decoupled learning of locomotion and manipulation behaviors, which improves scene generalization and cross-task transfer.

\section{MoMani}
\label{sec:benchmark}

\begin{figure*}[h] 
    \centering
    \includegraphics[width=0.75\textwidth]{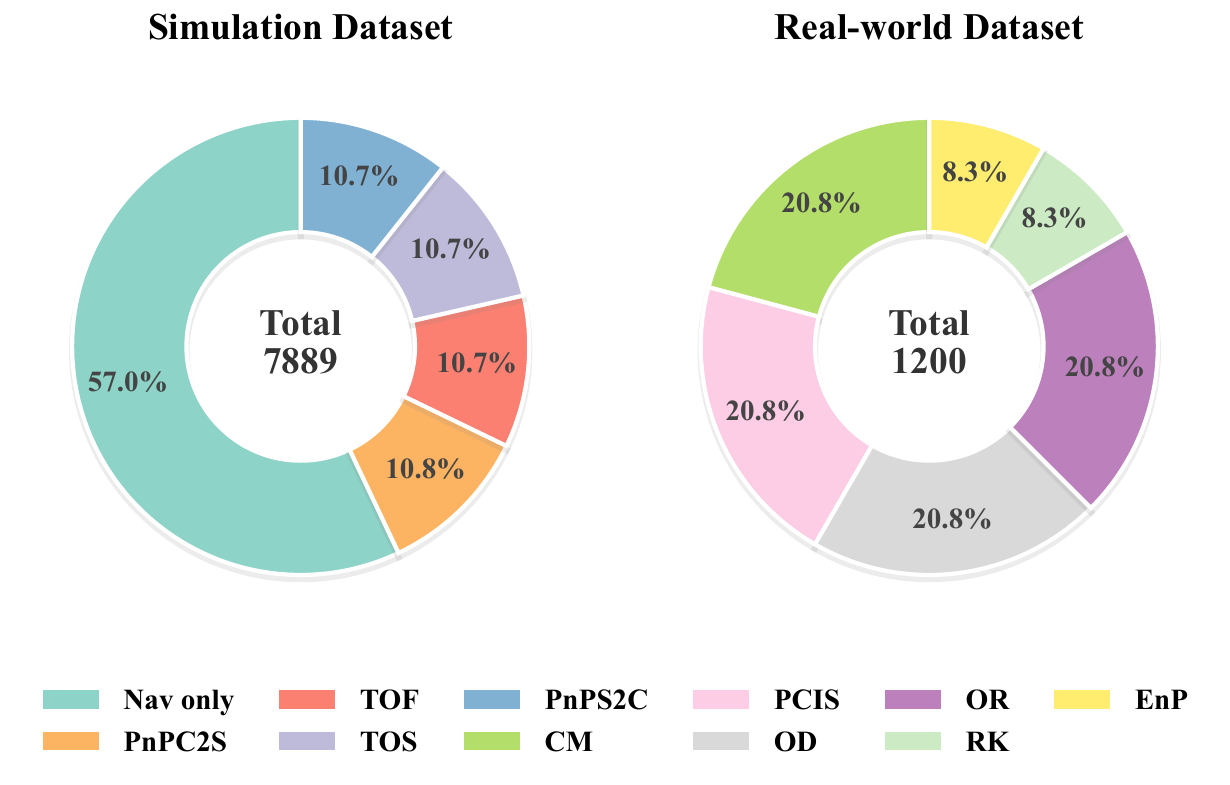} 
    \caption{Dataset composition. Simulation: pure navigation and four mobile manipulation tasks (TOF, PnPS2C, PnPC2S, TOS). Real-world: six mobile manipulation tasks (CM, OD, OR, PCIS, RK, EnP).}
    \label{fig:task_num}
\end{figure*}
We introduce \textbf{MoMani}, an automated benchmark for high-quality mobile manipulation data generation. As shown in Table~\ref{tab:comparison_momani}, while frameworks like ManiSkill2\cite{gu2023maniskill2} and RoboCasa\cite{nasiriany2024robocasa} focus on static tasks, MoMani provides a unified pipeline for integrated ``navigation + manipulation'' (Co-Gen) data.

Distinguishing itself from simulation-only platforms like BEHAVIOR-1K \cite{li2024behavior} and InfinitedWorld\cite{ren2024infiniteworld}, MoMani uniquely incorporates both simulated (S) and real-robot (R) data sources. It stands as the only benchmark to support a full spectrum of automated generation modes across both Mobile-M (Sim) and Real-Robot platforms, delivering expert-level trajectories for complex, multi-stage embodied tasks.

\begin{table*}[h]
\centering

\caption{Comparison of MoMani with representative embodied AI platforms. Sub-tasks indicates count per trajectory. Data Source: R (Real-world), S (Simulation); Generation: Navigation, Manipulation, Co-Gen; Robotic Platforms: Mobile-M (simulated), Real-Robot (physical); Action: N (Navigation), M (Manipulation).}
\label{tab:comparison_momani}
\resizebox{1\textwidth}{!}{
\begin{tabular}{l | c | cc | ccc | cc | c}
\toprule
\multirow{2}{*}{\textbf{Name}} & \multirow{2}{*}{\textbf{Sub-tasks}} & \multicolumn{2}{c|}{\textbf{Data Source}} & \multicolumn{3}{c|}{\textbf{Automated Generation}} & \multicolumn{2}{c|}{\textbf{Robotic Platforms}} & \multirow{2}{*}{\textbf{Action}} \\
& & Real (R) & Sim (S) & Navigation & Manipulation & Co-Gen & Mobile-M (Sim) & Real-Robot & \\
\midrule
ManiSkill2 \cite{gu2023maniskill2} & 1-2 & - & $\checkmark$ & - & $\checkmark$ & - & $\checkmark$ & - & \textit{M} \\
Social Nav \cite{puig2023habitat3} & 1 & - & $\checkmark$ & $\checkmark$ & - & - & $\checkmark$ & - & \textit{N, M} \\
HomeRobot \cite{yenamandra2023homerobot} & 3 & - & $\checkmark$ & $\checkmark$ & $\checkmark$ & - & $\checkmark$ & $\checkmark$ & \textit{N, M} \\
ProcTHOR \cite{deitke2022} & 3 & - & $\checkmark$ & $\checkmark$ & $\checkmark$ & - & $\checkmark$ & - & \textit{N, M} \\
Behavior-1K \cite{li2024behavior} & $>5$ & - & $\checkmark$ & $\checkmark$ & $\checkmark$ & - & $\checkmark$ & - & \textit{N, M} \\
Open X-Embodiment \cite{o2024open} & 1-3 & $\checkmark$ & - & - & - & - & $\checkmark$ & $\checkmark$ & \textit{M} \\
RoboCasa365 \cite{he2026nimbus} & 2-16 & - & $\checkmark$ & $\checkmark$ & $\checkmark$ & - & $\checkmark$ & - & \textit{N, M} \\
InfinitedWorld \cite{ren2024infiniteworld} & 3 & - & $\checkmark$ & $\checkmark$ & $\checkmark$ & $\checkmark$ & $\checkmark$ & - & \textit{N, M} \\

\midrule
\rowcolor[HTML]{F0F8EB}
\textbf{MoMani (Ours)} & 2-3 & $\checkmark$ & $\checkmark$ & $\checkmark$ & $\checkmark$ & $\checkmark$ & $\checkmark$ & $\checkmark$ & \textit{N, M} \\
\bottomrule
\end{tabular}
}
\end{table*}

\subsection{Simulation Data Construction}
\begin{figure*}[h]
    \centering
    \includegraphics[width=0.9\textwidth]{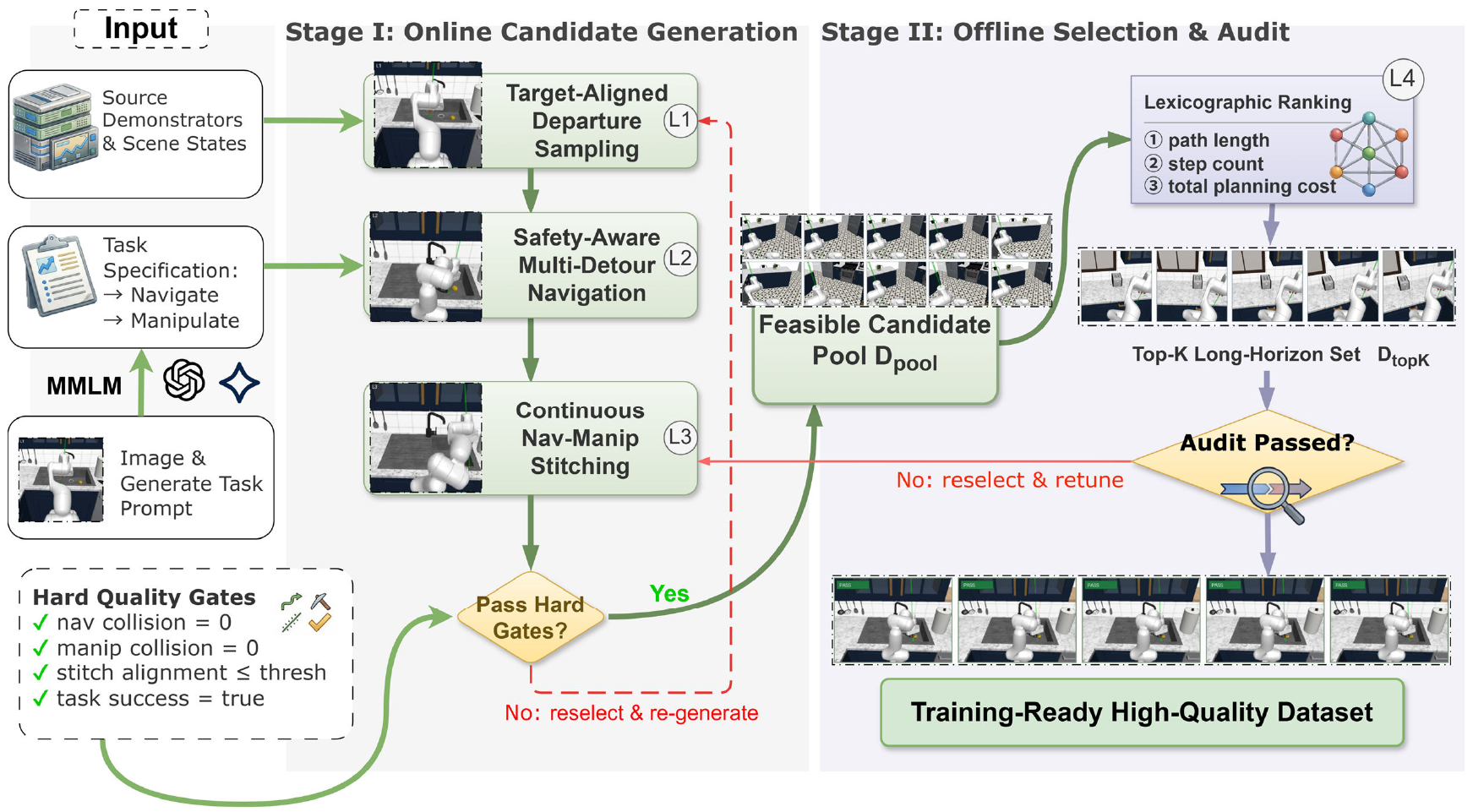}
   \caption{\textbf{Overview of the Quality-Controlled Data Engine.} Online generation synthesizes trajectories filtered by quality gates (collision-free, task success), while offline audit ranks candidates via lexicographic criteria (path length, cost) for training-ready data.}
    \label{fig:gen_pipeline}
\end{figure*}

As shown in Fig.~\ref{fig:gen_pipeline}, MoMani employs a two-stage automated pipeline to generate high-quality long-horizon trajectories. 

\textbf{Stage I: Online Candidate Generation.} Task specifications are processed via an MMLM prompter to execute sequential Target-Aligned Sampling (L1), Safety-Aware Navigation (L2), and Continuous Nav-Manip Stitching (L3). Crucially, our engine supports simultaneous base and arm execution, utilizing continuous collision detection to generate coordinated ``nav-manip'' trajectories. Candidates must pass Hard Quality Gates—verifying zero collisions, precise alignment ($\Delta \text{pos} < 0.05\text{m}, \Delta \text{ori} < 5^\circ$), and $100\%$ task success—before entering the feasible pool $D_{\text{pool}}$. 

\textbf{Stage II: Offline Selection \& Audit.} Candidates in $D_{\text{pool}}$ undergo Lexicographic Ranking (L4) based on path length and planning cost to identify the expert-level Top-K set $D_{\text{topK}}$. A final Scene-Camera Audit ensures the dataset is training-ready, enabling the scalable generation of $5,000+$ multimodal trajectories that bridge the gap between navigation and manipulation.

\subsection{Real-Robot Demonstration Collection}
To bridge the simulation-to-real gap, we collect a real-world dataset using a hardware configuration based on TidyBot++ \cite{wu2024tidybot++}. Our platform features a Kinova Gen3 7-DoF manipulator on a holonomic mobile base, equipped with front RGB-D and top-view stereo cameras synchronized via ROS. We utilize a web-based teleoperation interface to record multimodal data streams at 30 Hz, which are then segmented into motion primitives. Successful trajectories are verified via replay, while failed attempts are discarded.

\subsection{Dataset Distribution}
To provide a comprehensive evaluation for mobile manipulation, we construct a diverse and highly heterogeneous dataset that captures both broad environmental variations and intricate long-horizon task multi-modality. As visually summarized in Figure~\ref{fig:task_num}, our hierarchical data distribution comprises:
\begin{itemize}
    \item \textbf{Simulation (7,889 episodes):} Primarily consists of a large navigation-only subset (57.0\%) designed to prime base mobility, supplemented by diverse contact-rich mobile manipulation tasks including \textit{PnPC2S} (PickPlaceCounterToStove, 10.8\%), \textit{PnPS2C} (PickPlaceSinkToCounter, 10.7\%), \textit{TOS} (TurnOnStove, 10.7\%), and \textit{TOF} (TurnOnSinkFaucet, 10.7\%).
    \item \textbf{Real-world (1,200 episodes):} Features a balanced distribution (20.8\% each) of interactive manipulation tasks requiring precise contact, including \textit{OR} (Open refrigerator door), \textit{CM} (Close the microwave), \textit{OD} (Open the drawer), and \textit{PCIS} (Place the cup from the rack into the sink), along with additional long-horizon tasks such as \textit{EnP} (Enter the room and place the pears on the cabinet) and \textit{RK} (Rotate the knob of a microwave) (8.3\% each).
\end{itemize}

\section{Experiment}
\label{sec:experiment}

\begin{figure*}
    \centering
    \includegraphics[width=0.9\textwidth]{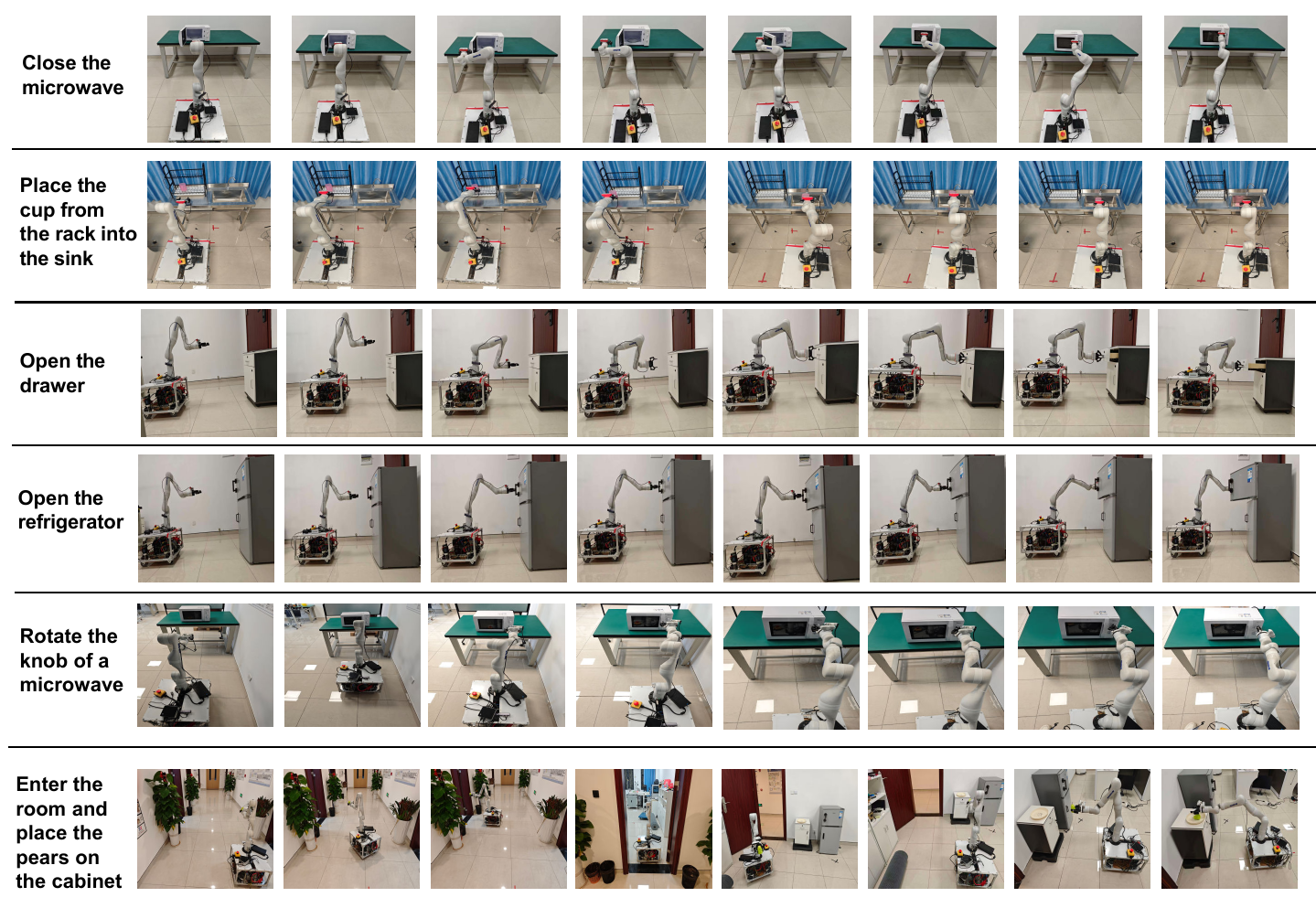}
    \caption{Real-world rollouts generated by EchoVLA. The robot successfully completes four mobile manipulation tasks—turning off the microwave, placing the cup from the rack into the sink, opening the drawer, and opening the refrigerator.}
    \label{fig:real_expr}
\end{figure*}


\subsection{Experimental Setup}
\label{sec:exp_setup}
We evaluate our proposed EchoVLA model in both simulation and real-world environments. 
In \textbf{simulation}, experiments are conducted in the RoboCasa simulator using multiple mobile manipulation tasks. 
For \textbf{real-world experiments}, we deploy EchoVLA using the \textit{TidyBot++} platform \cite{wu2024tidybot++} in a $7\text{m} \times 7\text{m}$ arena to execute diverse household tasks, notably \textit{EnP} (Enter and Pick). 

Unlike other stationary or single-room tasks, \textit{EnP} requires cross-room navigation and manipulation, moving from a starting corridor into a target room to retrieve objects. This validates the model's capability in handling long-horizon spatial transitions and multi-stage logic in complex environments.

We train EchoVLA on 8 NVIDIA A100 GPUs with observations including multi-view RGB-D images and robot states. And the action space is decomposed into mobile base and arm actions following the per-part diffusion design.

\begin{table*}[t]
\fontsize{8}{8}\selectfont
\centering
\renewcommand{\arraystretch}{1.3} 
\setlength{\tabcolsep}{6pt} 

\definecolor{headergray}{RGB}{240, 240, 240}
\definecolor{oursgreen}{RGB}{235, 245, 235}
\caption{
Success Rate (SR) comparison on Manipulation, Navigation, and Mobile Manipulation tasks. 
EchoVLA (Ours) achieves the highest performance in most scenarios.
}

\begin{tabularx}{\textwidth}{l l *{6}{>{\centering\arraybackslash}X}}
\toprule
\rowcolor{headergray}
\textbf{Category} & \textbf{Method} & \textbf{PnP C2S} & \textbf{PnP S2C} & \textbf{TOF} & \textbf{TOS} & \textbf{Nav only} & \textbf{Avg SR} \\
\midrule

\multirow{6.5}{*}{\rotatebox[origin=c]{90}{\shortstack{\textbf{Manip. /} \\ \textbf{Nav.}}}} 
 & BC-T & 0.04 & 0.46 & 0.33 & 0.45 & 0.10 & 0.28 \\
 & Diffusion Policy & 0.00 & 0.00 & 0.03 & 0.00 & 0.03 & 0.01 \\
 & DP3 & 0.03 & 0.25 & 0.43 & 0.26 & 0.13 & 0.22 \\
 & WB-VIMA & 0.00 & 0.12 & 0.11 & 0.19 & 0.31 & 0.15 \\
 & $\boldsymbol{\pi}_{0.5}$ & 0.18 & 0.20 & 0.44 & 0.52 & 0.24 & 0.32 \\
\cmidrule(lr){2-8} 
\rowcolor{oursgreen}
 & \textbf{EchoVLA (Ours)} & \textbf{0.21} & \textbf{0.68} & \textbf{0.51} & \textbf{0.67} & \textbf{0.51} & \textbf{0.52} \\

\midrule 

\multirow{5.5}{*}{\rotatebox[origin=c]{90}{\shortstack{\textbf{Mobile} \\ \textbf{Manip.}}}} 
 & BC-T & 0.00 & 0.11 & 0.04 & 0.04 & -- & 0.05 \\
 & DP3 & 0.00 & 0.10 & 0.08 & 0.04 & -- & 0.06 \\
 & WB-VIMA & 0.00 & 0.12 & 0.11 & 0.19 & -- & 0.11 \\
 & $\boldsymbol{\pi}_{0.5}$ & 0.08 & 0.25 & 0.18 & 0.27 & -- & 0.20 \\
\cmidrule(lr){2-8}
\rowcolor{oursgreen}
 & \textbf{EchoVLA (Ours)} & \textbf{0.17} & \textbf{0.34} & \textbf{0.29} & \textbf{0.43} & -- & \textbf{0.31} \\

\bottomrule
\end{tabularx}
\label{tab:manip_nav}
\end{table*}

\subsection{Comparison with Baselines in RoboCasa Simulation}
\label{sec:exp_baseline}

We evaluate \textbf{EchoVLA} 
against representative baselines  
including 
BC-T~\cite{bct}, Diffusion Policy~\cite{chi2024diffusionpolicy}, DP3~\cite{yang2024diffusion3d}, $\boldsymbol{\pi}_{0.5}$~\cite{pi052025}, and WB-VIMA~\cite{jiang2025behaviorrobotsuitestreamlining}.
The evaluation spans four core tasks in the RoboCasa simulator: 
PickPlaceCounterToStove, PickPlaceSinkToCounter, TurnOnSinkFaucet, and TurnOnStove.
To assess \textbf{Mobile Manipulation}, we introduce additional challenges by initializing the robot at a distance from targets or requiring follow-up navigation post-manipulation. 

Performance is analyzed across three categories: (1) \textbf{Manipulation} (arm-only), (2) \textbf{Navigation} (base precision), and (3) \textbf{Mobile Manipulation} (coordinated control). We report the \textbf{Success Rate (SR)} averaged over three random seeds and 50 evaluation episodes per task. These findings are consolidated in Table~\ref{tab:manip_nav}, highlighting the inherent difficulty of coordinated mobile manipulation. The results in Table~\ref{tab:manip_nav} reveal several important observations:

\textbf{1) Method Comparison.} 
For Manipulation and Navigation tasks, traditional behavior cloning (BC-T) and Diffusion Policy exhibit low performance, with Diffusion Policy often achieving near-zero success rates. $\boldsymbol{\pi}_{0.5}$ shows improved performance, while our proposed EchoVLA achieves the highest average success rate (Avg Success = 0.52), demonstrating robust task execution.

\textbf{2) Increased Difficulty in Mobile Manipulation.} 
Coordinating base and arm control significantly increases complexity, causing success rates to drop across all methods. For instance, $\pi_{0.5}$'s average success decreases from 0.32 to 0.20, while BC-T exhibits a drastic decline from 0.28 to 0.05.

\textbf{3) Superiority of EchoVLA.} 
Despite the increased difficulty, EchoVLA maintains the highest success rate (0.31) in Mobile Manipulation tasks, outperforming all other methods. This highlights the method's effectiveness in handling coordinated base–arm motion, long-range navigation, and sequential manipulation tasks.

\textbf{4) Task-specific Performance Variations.} 
Tasks such as 
PickPlaceSinkToCounter
and 
TurnOnSinkFaucet
show larger performance variations across methods, reflecting higher requirements for spatial understanding and precise manipulation. EchoVLA consistently achieves strong performance on these challenging tasks, demonstrating its capability in multi-modal perception and action reasoning.

\subsection{Ablation Study}
\label{sec:exp_ablation}
Ablations are performed on two representative RoboCasa tasks, \textbf{PnP Counter to Stove (Mobile)} and \textbf{PnP Counter to Stove}, and results are reported as \textbf{Success Rate (SR)} averaged over 50 episodes.

\begin{table}[t]
    \centering
    \footnotesize 
    \setlength{\tabcolsep}{12pt} 
    \renewcommand{\arraystretch}{1.1}
    \caption{Ablation results on the \textbf{PnP Counter to Stove} task. M: Mobile; S: Static.}
    \label{tab:ablation_obs_memory}
    \begin{tabular}{l cccc | cc} 
        \toprule
        & \textbf{RGB} & \textbf{PC} & \textbf{EM} & \textbf{SM} & \textbf{M} & \textbf{S} \\ 
        \midrule
        (a) & \ding{55} & \checkmark & \checkmark & \checkmark & 0.02 & 0.13 \\
        (b) & \checkmark & \ding{55} & \checkmark & \checkmark & 0.08 & 0.15 \\
        (c) & \checkmark & \checkmark & \checkmark & \ding{55} & 0.09 & 0.16 \\
        (d) & \checkmark & \checkmark & \ding{55} & \checkmark & 0.14 & 0.13 \\
        \midrule
        \rowcolor[RGB]{240,248,235}
        \textbf{Ours} & \checkmark & \checkmark & \checkmark & \checkmark & \textbf{0.17} & \textbf{0.21} \\
        \bottomrule
    \end{tabular}
\end{table}

\textbf{1) Observation Ablation.}
We examine the effect of removing point cloud input from the observation space. When the agent relies solely on RGB features without 3D geometric cues, its spatial grounding ability decreases notably. As shown in Table~\ref{tab:ablation_obs_memory}, removing point cloud input results in consistent performance drops across both mobile and non-mobile variants, demonstrating the importance of geometric depth information for accurate object localization and grasping.

\textbf{2) Memory Module Ablation.}
We further analyze the roles of our two memory mechanisms: \textbf{Episodic Memory (EM)} and \textbf{Scene Memory (SM)}.
The ablation rows in Table~\ref{tab:ablation_obs_memory} show that disabling either EM or SM leads to substantial degradation, and removing both RGB or PC signals further compounds these effects. Overall, the results highlight that observation diversity (RGB + PC) and hierarchical memory (EM + SM) are both essential for robust performance in manipulation and mobile manipulation tasks in RoboCasa.




\begin{table}[h] 
    \centering
    \footnotesize
    \definecolor{headercolor}{RGB}{245, 245, 245}
    \setlength{\tabcolsep}{4pt} 
    \renewcommand{\arraystretch}{1.2}
    \caption{Sensitivity analysis of hyperparameters on the PnPC2S (Mobile) task. $L=8$ and $\tau=0.5$ are selected as optimal values for EchoVLA.}
    \label{tab:sensitivity_combined}

    \begin{tabularx}{0.7\linewidth}{X p{25pt} p{25pt} p{25pt} p{25pt}}
        \toprule
        \rowcolor{headercolor}
        \textbf{Window Size ($L$)} & \centering 2 & \centering 4 & \centering 8 & \centering 16 \tabularnewline
        \midrule
        SR $\uparrow$ & \centering 0.08 & \centering 0.12 & \centering \textbf{0.17} & \centering 0.15 \tabularnewline
        \bottomrule
    \end{tabularx}

    \vspace{1.5mm} 

    \begin{tabularx}{0.7\linewidth}{X p{25pt} p{25pt} p{25pt} p{25pt}}
        \toprule
        \rowcolor{headercolor}
        \textbf{Threshold ($\tau$)} & \centering 0.1 & \centering 0.3 & \centering 0.5 & \centering 0.7 \tabularnewline
        \midrule
        SR $\uparrow$ & \centering 0.11 & \centering 0.14 & \centering \textbf{0.17} & \centering 0.13 \tabularnewline
        \bottomrule
    \end{tabularx}
\end{table}
\textbf{3) Sensitivity Analysis.} We evaluate EchoVLA's sensitivity to EM window size $L$ and update threshold $\tau$. As shown in Table~\ref{tab:sensitivity_combined}, a small $L$ (2 or 4) leads to ``plan forgetting'' and oscillations, while $L=16$ introduces latency with diminishing returns. $L=8$ provides the optimal performance-efficiency balance. Furthermore, a low threshold ($\tau=0.1$) causes representation instability, whereas a high threshold ($\tau=0.7$) results in outdated scene memory. $\tau=0.5$ strikes the best balance, ensuring memory freshness and stable spatial grounding for long-horizon tasks.

\subsection{Real Robot Experiment}
We evaluate EchoVLA across a diverse set of tasks, including opening a drawer (\textit{OD}), closing a microwave (\textit{CM}), placing a cup into the sink (\textit{PCIS}), opening a refrigerator (\textit{OR}), rotating a knob (\textit{RK}), and entering a room to place pears (\textit{EnP}). For each task, we conduct 20 independent trials with randomized initial robot base positions to ensure statistical significance.
\label{sec:exp_real}
\textbf{1) Real-World Performance.} As shown in Table~\ref{tab:real_robot_mobile}, EchoVLA achieves a 0.44 mean success rate, outperforming $\pi_{0.5}$ (0.33) and Diffusion Policy (0.32). While baselines suffice for shorter tasks like \textit{OD} and \textit{OR}, EchoVLA excels in complex ones like \textit{CM} (0.70) and \textit{RK} (0.50). Crucially, it leads in the longest-horizon task, \textit{EnP} (362.9 frames), confirming that our synergistic memory stabilizes control and mitigates perceptual noise.
\begin{table*}[h]
\centering
\small 
\renewcommand{\arraystretch}{1.2} 
\definecolor{headercolor}{RGB}{245, 245, 245}
\definecolor{oursrow}{RGB}{235, 248, 235} 
\setlength{\tabcolsep}{4pt} 
\caption{Success rates for real robot mobile manipulation tasks. We report performance on six basic actions. Avg SR is the mean success rate. (\textbf{Notes:} OD: Open the drawer, CM: Close the microwave, PCIS: Place the cup from the rack into the sink, OR: Open refrigerator door, RK: Rotate the knob of a microwave, EnP: Enter the room and place the pears on the cabinet.)}

\begin{tabularx}{\textwidth}{l *{6}{>{\centering\arraybackslash}X} >{\centering\arraybackslash}p{2.2cm}} 
\toprule
\rowcolor{headercolor}
\textbf{Method} & \textbf{OD} & \textbf{CM} & \textbf{PCIS} & \textbf{OR} & \textbf{RK} & \textbf{EnP} & \textbf{Avg SR} \\
\midrule
$\boldsymbol{\pi}_{0.5}$ & 0.35 & 0.55 & 0.20 & 0.50 & 0.40 & 0.00 & 0.33 \\
Diffusion Policy& 0.40 & 0.60 & 0.50 & 0.30 & 0.10 & 0.03 & 0.32 \\
\rowcolor{oursrow}
\textbf{EchoVLA (Ours)} & \textbf{0.45} & \textbf{0.70} & \textbf{0.50} & \textbf{0.40} & \textbf{0.50} & \textbf{0.10} & \textbf{0.44} \\
\bottomrule
\end{tabularx}
\label{tab:real_robot_mobile}
\end{table*}

\textbf{2) Robustness to Real-World Perceptual Noise.} 
As shown in Table~\ref{tab:real_robot_mobile}, EchoVLA achieves a mean success rate of 0.44, surpassing Diffusion Policy (0.32). EchoVLA excels in long-horizon precision, achieving 50\% SR on \textit{RK} while baselines fail (10\%) due to memory drift. This gap suggests that temporal context from EM acts as a corrective anchor, compensating for spatial inaccuracies like voxel map ``ghosting'' in scene memory.

\textbf{3) Failure Analysis on Dynamic Occlusion.} 
Despite its robustness, performance drops in \textit{OR} due to dynamic occlusion. As the fridge door opens, rapidly changing geometry degrades EchoVLA's explicit 3D Scene Memory. Conversely, baselines relying on implicit ``muscle memory'' are more robust to such blind spots where explicit geometry fails. This confirms that while synergistic memory ensures overall stability, explicit 3D representations remain sensitive to extreme structural dynamics.

\section{Conclusion}
\label{sec:conclusion}
We present EchoVLA, a memory-aware vision–language–action framework for mobile manipulation. By integrating a 3D Scene Memory with an Episodic Memory, EchoVLA effectively handles non-Markovian decision-making and mitigates perceptual noise. To validate our approach, we introduce MoMani, a benchmark spanning procedurally generated simulation tasks and real-robot evaluations. Extensive experiments demonstrate that EchoVLA consistently outperforms strong baselines across diverse scenarios.

\noindent\textbf{Limitations \& Future Work.} 
EchoVLA's performance depends on high-quality depth and pose streams; cumulative odometry drift can cause spatial misalignments or ``ghosting'' in the Scene Memory. Future work will integrate loop-closure or visual-SLAM to improve tracking robustness. Furthermore, to address the cold-start problem in novel environments where the 3D memory is initially sparse, we plan to explore active exploration strategies or generative 3D priors. Lastly, we will investigate alternative encoder architectures—such as unified multi-modal transformers—to enhance scalability and cross-task generalization.


\section*{Acknowledgements}
This work is supported by the National Natural Science Foundation of China (NSFC) under Grant No. 62506231, the National Key Research and Development Program of China (2024YFE0203100), the \mbox{Scientific} Research Innovation \mbox{Capability} Support Project for Young Faculty (No. ZYGXQNJSKYCXNLZCXM\allowbreak-I28), the Shenzhen Science and Technology Program (Grant No. QNXMA20250701093544048), and the General Embodied AI Center of Sun Yat-sen University.
%
%
\bibliographystyle{splncs04}
\bibliography{main}
\end{document}